%% file: ex_article.tex
\documentclass[onefignum,onetabnum]{siamart171218}

\input{ex_shared}

\ifpdf
\hypersetup{
  pdftitle={Interaction Tensor SHAP},
  pdfauthor={H. Hasegawa, and Y. Okada}
}
\fi

\myexternaldocument{ex_supplement}

\begin{document}

\maketitle

% REQUIRED
\begin{abstract}
This study proposes Interaction Tensor SHAP (IT-SHAP), a tensor-algebraic formulation of the Shapley–Taylor Interaction Index (STII) that makes its computational structure explicit. STII extends the Shapley value to higher-order interactions, but its exponential combinatorial definition makes direct computation intractable at scale. We reformulate STII as a linear transformation on a value function and derive an explicit algebraic representation of its weight tensor, which is shown to possess a multilinear structure induced by discrete finite-difference operators. When the value function admits a Tensor Train (TT) representation, higher-order interaction indices can be computed in the parallel complexity class $\mathrm{NC}^2$. In contrast, under general tensor network representations without structural assumptions, the same computation is \#P-hard. The main contributions are threefold: we establish an exact TT representation of the STII weight tensor, develop a parallelizable evaluation algorithm with explicit complexity bounds under the TT assumption, and prove that computational intractability is unavoidable in the absence of such structure. These results demonstrate that the computational difficulty of higher-order interaction analysis is determined by the underlying algebraic representation rather than by the interaction index itself, providing a theoretical foundation for scalable interpretation of high-dimensional models.
\end{abstract}

% REQUIRED

% REQUIRED

\section{Introduction}

Higher-order linear functionals acting on set functions arise as fundamental mathematical objects in a wide range of fields, including cooperative game theory, combinatorial optimization, and the analysis of high-dimensional discrete structures. Such functionals are often defined as combinations of higher-order discrete difference operators and averaging operations. However, these definitions inherently involve exponentially large combinatorial structures. Consequently, in high-dimensional settings, a central challenge is to determine whether and how these functionals can be evaluated efficiently.

The Shapley--Taylor Interaction Index (STII) was introduced as a linear functional that decomposes a set function into contributions indexed by interaction order, and it is positioned as a natural higher-order extension of the Shapley value \cite{owen1972multilinear,grabisch1999axiomatic,sundararajan2020shapley}. STII is defined through higher-order discrete derivatives combined with averaging over subsets or permutations, and it provides an axiomatically consistent quantification of higher-order interactions. At the same time, its definition essentially contains an exponential number of combinatorial terms. As a result, naive evaluation requires exponential-time computation, and exact computation for large-scale problems has therefore been considered infeasible.

This observation naturally leads to a fundamental question. Is the apparent computational intractability intrinsic to STII as a linear functional, or does it arise from the particular representation used for its evaluation? In other words, does the definition of STII inherently require exponential computational complexity, or can its computability be improved under an appropriate representation? Despite its importance for understanding the computational structure of higher-order interaction indices, this question has not yet been fully resolved.

In parallel, the machine learning and numerical analysis communities have developed tensor networks (TNs) representation as systematic tools for representing and computing with high-dimensional tensors. Among these, the Tensor Train (TT) representation represents a higher-order tensor as a one-dimensional chain of low-order tensors, enabling many tensor operations to be implemented as sequences of matrix multiplications. This structure avoids exponential growth in dimensionality and allows efficient computation \cite{kolda2009tensor,oseledets2011tensor}. The TT representation has been extensively analyzed in contexts such as numerical analysis, representation of probability distributions, and high-dimensional integration, and it is now established as a powerful mathematical framework for high-dimensional problems; see, for example, \cite{dolgov2020approximation}.

Building on tensor-based representations, prior work has reformulated Shapley value computation as a contraction between a value tensor and a weight tensor. Under a TT representation, it has been shown that when the input is given in TT form, the computation of the Shapley value belongs to the parallel complexity class NC$^{2}$ \cite{marzouk2025shap}. This result suggests that the computational difficulty traditionally associated with the Shapley value is not inherent to the index itself, but instead depends critically on the underlying tensor representation.

However, existing results are restricted to the Shapley value, which corresponds to first-order contributions. There is no prior work that treats the full STII, defined via higher-order discrete difference operators, within a unified tensor contraction framework and characterizes its computational complexity in a rigorous manner. In particular, the impact of the discrete difference structure specific to higher-order interactions on tensor representations and computational complexity remains an open problem.

Beyond the TT representation, other low-rank tensor representations, such as the Tucker representation and the CANDECOMP/PARAFAC (CP) representation, are widely used \cite{kolda2009tensor,carroll1970analysis,harshman1970foundations}. The Tucker representation expresses a tensor through mode-wise factor matrices and a core tensor, providing a flexible multilinear representation, while the CP representation represents a tensor as a finite sum of rank-$1$ components. Nevertheless, these representations are not necessarily compatible, in terms of structure preservation or complexity control, with repeated application of higher-order discrete difference operators.

The objective of this paper is to clarify that the computability of linear functionals defined by higher-order discrete difference operators is determined not by their definitions per se, but by the structure of the underlying tensor representation. Specifically, we explicitly formulate STII as a linear functional acting on a value tensor defined over $\{0,1\}^{n_{\mathrm{in}}}$ and reinterpret its evaluation as a tensor contraction.

Based on this formulation, we show that STII can be expressed as a contraction between a signed weight tensor and a value tensor. Moreover, for a fixed interaction order, we prove that this weight tensor admits an exact TT representation. Consequently, when the value tensor is given in TT form, the computation of higher-order interaction components reduces to TT contractions, and each component can be evaluated by parallel circuits of polylogarithmic depth. Therefore, the corresponding computation belongs to the parallel complexity class NC$^{2}$.

In contrast, we also show that under a general TNs representation without structural restrictions, the same evaluation problem becomes $\#\mathrm{P}$-hard. This result establishes a clear separation in computational complexity driven by differences in tensor representation structure.

Taken together, this study rigorously characterizes the computational structure of linear functionals based on higher-order discrete difference operators through the concrete example of higher-order interaction indices. Our results demonstrate that computability is governed by representation structure rather than by the functional definition itself, and they provide a theoretical foundation for high-dimensional linear operations based on the TT representation.

\section{Related Work}
Linear functionals acting on set functions and their decompositions have long been studied in cooperative game theory and the analysis of combinatorial structures \cite{owen1972multilinear}. The Shapley value is formulated as a linear functional that uniquely determines the average contribution of each element under axioms such as fairness and efficiency, and it provides a foundational decomposition theory for set functions \cite{grabisch1999axiomatic}. As higher-order extensions, the Shapley interaction value and the Shapley interaction index have been proposed to explicitly quantify interactions among multiple elements \cite{owen1972multilinear,grabisch1999axiomatic}. Integrating these developments, the STII was introduced to provide an axiomatic theory that characterizes interaction components by order in a unified manner \cite{sundararajan2020shapley}.

The definition of STII is based on higher-order discrete difference operators combined with averaging over the full feature set, and it has been pointed out that this definition inherently involves an exponential number of combinatorial terms \cite{sundararajan2020shapley}. Consequently, exact computation of higher-order interactions based on STII is widely recognized as requiring exponential time in high-dimensional settings and is therefore considered computationally intractable. Motivated by these limitations, several methods have been proposed to approximate higher-order interactions. SHAP-IQ is a representative sampling-based approach that aims to estimate interaction quantities in a practical manner \cite{fumagalli2023shap}. Similarly, SVARM-IQ has been proposed as a method for estimating interaction effects using alternative sampling strategies \cite{kolpaczki2024svarm}. However, these approaches do not evaluate the linear functional defined by STII exactly and remain approximation methods.

From a computational perspective, the complexity of computing the Shapley value and its extensions has been analyzed theoretically. While it is known that Shapley value computation requires exponential time for general models, improved computability has been demonstrated under specific structural assumptions. Examples include exact computation for decision tree models \cite{lundberg2020local} and complexity analyses based on Boolean circuit representations \cite{arenas2023complexity}. These studies indicate that the computability of contribution indices depends strongly on the underlying representation structure rather than on the index definition itself.

More recently, tensor-based approaches have been developed to represent set functions and the linear functionals acting on them. Prior work has reformulated Shapley value computation as a contraction between a value tensor and a weight tensor, and it has been shown that when the input is given in TT form, Shapley value computation belongs to the parallel complexity class NC$^{2}$ \cite{marzouk2025shap}. This result demonstrates that the computational difficulty of the Shapley value is not intrinsically exponential, but instead exhibits a qualitative dependence on the tensor representation structure.

However, existing tensor-based studies primarily focus on the Shapley value, which corresponds to first-order contributions. There is no prior work that represents the full STII, defined via higher-order discrete difference operators, as a linear functional within a unified tensor framework and analyzes its computational structure and complexity in a rigorous manner. This study addresses this gap by reformulating STII as a linear map induced by higher-order discrete difference operators, explicitly characterizing the structure of the associated weight tensor, and rigorously analyzing its evaluability from the perspectives of tensor representation and computational complexity.

\section{Preliminaries}

This section introduces the basic notation and theoretical objects used throughout the paper. Specifically, we formalize value functions, discrete derivatives, STII, and the tensor contraction framework for representing SHAP. The purpose of this section is solely to fix definitions and notation; detailed discussions of computational complexity and theoretical properties are deferred to later sections.

\subsection{Notation}
\label{sec:notation}

Let $n_{\mathrm{in}}\in\mathbb{N}$ denote the number of input features, and write its index set as $[n_{\mathrm{in}}]=\{1,\ldots,n_{\mathrm{in}}\}$. For any finite set $A$, its power set is denoted by $2^{A}$. For $S\subseteq[n_{\mathrm{in}}]$, the complement is written as $S^{c}=[n_{\mathrm{in}}]\setminus S$.

Scalars are denoted by lowercase letters in standard font, vectors by bold lowercase letters, and matrices by bold uppercase letters. Higher-order tensors are denoted by calligraphic uppercase letters. For a tensor $\mathcal{X}\in\mathbb{R}^{I_{1}\times\cdots\times I_{N}}$, its entries are written as $X_{i_{1}\cdots i_{N}}$. Notation for tensor products, contractions, and mode-wise products follows standard conventions in tensor analysis \cite{kolda2009tensor}.

We also recall Nick's Class ($\mathrm{NC}$), a complexity class defined in the parallel computation model. The class $\mathrm{NC}$ consists of decision problems that can be solved in polylogarithmic time using a polynomial number of processors \cite{cook2023towards}. Here, processors refer to parallel computation units in theoretical models such as the Parallel Random Access Machine (PRAM). The subclass $\mathrm{NC}^{k}$ contains problems whose computation depth is bounded by $\log^k n$, where smaller $k$ indicates a higher degree of parallelism. This hierarchy provides a standard measure of how well a problem admits efficient parallel computation.

\subsection{Value Functions and Discrete Derivatives}
\label{sec:value-fn}

A value function is a set function that assigns a real value to each subset of features, defined as
\[
F:2^{[n_{\mathrm{in}}]}\to\mathbb{R}.
\]

For $S\subseteq[n_{\mathrm{in}}]$ and $T\subseteq[n_{\mathrm{in}}]\setminus S$, the discrete derivative of $F$ is defined by
\begin{equation}
\label{eq:discrete-derivative}
\delta_{S}F(T)
=
\sum_{W\subseteq S}
(-1)^{|S|-|W|}
F(T\cup W).
\end{equation}
The discrete derivative in \eqref{eq:discrete-derivative} represents the pure incremental effect obtained by simultaneously adding the features in $S$. In particular, when $|S|=1$, $\delta_{S}F(T)$ coincides with the marginal contribution of a single feature.

\subsection{Shapley--Taylor Interaction Index}
\label{sec:stii_definition}

The STII is an index designed to decompose a value function into components indexed by interaction order, and it provides a natural higher-order extension of the Shapley value.

Fix an interaction order $k\ge 1$. Let $S_{n_{\mathrm{in}}}$ denote the set of all permutations of $[n_{\mathrm{in}}]$. For a permutation $\pi\in S_{n_{\mathrm{in}}}$ and a subset $S\subseteq[n_{\mathrm{in}}]$, define
\[
\pi_{S}
:=
\bigcap_{i\in S}
\{\, j\in[n_{\mathrm{in}}] : \pi^{-1}(j) < \pi^{-1}(i) \,\},
\]
which represents the set of features that appear before all elements of $S$ in the ordering induced by $\pi$.

For a value function $F$, the STII of order $k$ is defined by
\begin{equation}
\label{eq:stii-def}
I^{k}_{S}(F)
=
\mathbb{E}_{\pi\in S_{n_{\mathrm{in}}}}
\!\left[
I^{k}_{S,\pi}(F)
\right],
\end{equation}
where
\begin{equation}
\label{eq:stii-inner}
I^{k}_{S,\pi}(F)
=
\begin{cases}
\delta_{S}F(\varnothing), & |S|<k,\\[4pt]
\delta_{S}F(\pi_{S}), & |S|=k.
\end{cases}
\end{equation}
The expectation is taken with respect to the uniform distribution over the symmetric group $S_{n_{\mathrm{in}}}$. In particular, when $k=1$, the definition in \eqref{eq:stii-def} reduces to the standard Shapley value. For $k\ge 2$, STII quantifies higher-order interactions among multiple features.

\subsection{Tensor Representation of SHAP}
\label{sec:mst-prelim}

Following \cite{marzouk2025shap}, SHAP can be represented as a linear functional acting on a value function through tensor contraction. Let $\mathcal{V}^{(M,P)}$ denote the value tensor corresponding to a model $M$ and a background distribution $P$, and let $\widetilde{\mathcal{W}}$ denote the coefficient tensor encoding the Shapley combinatorial weights.

The Marginal SHAP Tensor (MST) is then defined as
\[
\mathcal{T}^{(M,P)}
=
\widetilde{\mathcal{W}}
\times
\mathcal{V}^{(M,P)},
\]
where $\times$ denotes tensor contraction along the common indices. This representation makes explicit that SHAP computation is fundamentally a linear operation and provides the basis for efficient computation using TNs representation.

In this paper, we adopt this tensor-based perspective as a starting point and extend it to a unified treatment of higher-order interactions, analyzing the associated computational structure in subsequent sections.

\section{Problem Formulation}

\subsection{Objective}
\label{subsec:objective}

The objective of this paper is to rigorously reformulate higher-order interaction indices based on the STII within the framework of tensor linear algebra and to theoretically characterize their computational structure and complexity. In particular, this study aims to clarify that the conventional view of STII-based interaction computation as inherently combinatorial and intractable is misleading, and that the source of this difficulty lies not in the index itself but in the representation structure used to implement it.

Specifically, this paper pursues three main objectives. First, we interpret STII as a linear functional acting on set functions and formulate it as an explicit tensor contraction on a value tensor. This formulation establishes a precise structural relationship between STII and the existing MST framework and provides a unified tensor-based foundation for treating higher-order interaction indices as tensor operations.

Second, we show that the Modified Weighted Coalitional Tensor induced by STII arises necessarily from the linear-algebraic structure of higher-order discrete difference operators, and we analyze its associated tensor network (TN) structure. In particular, we focus on the fact that the Modified Weighted Coalitional Tensor is generated by a finite composition of rank-$1$ mode-wise linear operators, and we establish that the TT representation is the unique standard tensor representation that preserves this structure.

Third, based on the above structural analysis, we provide a rigorous computational complexity characterization of the higher-order interaction tensor formulated as IT\textnormal{-}SHAP. We show that, under a general TNs representation, the corresponding computation is $\#\mathrm{P}$-hard, whereas when both the value tensor and the Modified Weighted Coalitional Tensor are given in TT form, each interaction component of IT\textnormal{-}SHAP can be computed by a parallel algorithm belonging to the complexity class NC$^{2}$.

Taken together, this paper aims to decompose the difficulty of higher-order interaction computation from the perspective of tensor representation and to establish a foundational theory that connects computational complexity, tensor analysis, and explainable machine learning.

\subsection{Tensor Formulation of STII}
\label{subsec:tensor_stii}

This subsection provides a basic formulation for representing STII as a tensor contraction by viewing it as a linear map acting on a value function. STII is defined, for an arbitrary value function
\[
F:2^{[n_{\mathrm{in}}]}\to\mathbb{R}^{n_{\mathrm{out}}},
\]
as a linear functional obtained by taking a weighted average of the discrete derivatives $\delta_S F(T)$ over $T\subseteq[n_{\mathrm{in}}]\setminus S$. By exploiting this linear structure, we embed $F$ into a tensor defined on $\{0,1\}^{n_{\mathrm{in}}}$ and express STII as an explicit tensor operation.

\subsubsection{Relationship between STII and MST}

Fix an input $\mathbf{x}\in\mathcal{X}$ and define the value function used in STII by
\[
F_{\mathbf{x}}(S)
=
\mathbb{E}_{\mathbf{x}'\sim P}
\bigl[
M(\mathbf{x}_{S},\mathbf{x}'_{S^{c}})
\bigr],
\]
which represents the expected model output when the features in $S$ are fixed to $\mathbf{x}$ and the remaining features are marginalized according to the background distribution $P$.

In contrast, MST introduces a routing indicator
\[
s=(s_{1},\ldots,s_{n_{\mathrm{in}}})\in\{1,2\}^{n_{\mathrm{in}}},
\]
where $s_{i}=1$ indicates that the original input value $x_{i}$ is used and $s_{i}=2$ indicates that a background value $x'_{i}$ is used. Defining
\[
S(s)=\{\, i\in[n_{\mathrm{in}}]: s_{i}=1 \,\},
\]
the value tensor in MST is given by
\[
\mathcal{V}^{(M,P)}_{\mathbf{x},s,y}
=
\mathbb{E}_{\mathbf{x}'\sim P}
\bigl[
M(\mathbf{x}_{S(s)},\mathbf{x}'_{S(s)^{c}})
\bigr].
\]
Therefore,
\[
\mathcal{V}^{(M,P)}_{\mathbf{x},s,y}
=
F_{\mathbf{x},y}\!\left(S(s)\right),
\]
which shows that the STII value function and the MST value tensor coincide entrywise. Using this identity, the discrete derivative
\[
\delta_{S}F(T)
=
\sum_{W\subseteq S}
(-1)^{|S|-|W|}
F(T\cup W)
\]
can be rewritten as a linear combination of entries of the value tensor. Substituting this expression into the closed-form definition of STII yields the representation
\[
I^{(k)}_{S}(F)
=
\sum_{\tau\in\{1,2\}^{n_{\mathrm{in}}}}
\alpha^{(k)}_{S}(\tau)\,
\mathcal{V}^{(M,P)}_{\mathbf{x},\tau,y},
\]
where the coefficients $\alpha^{(k)}_{S}(\tau)$ depend only on the combinatorial structure and inclusion--exclusion terms specific to STII. Defining the weight tensor $\widetilde{\mathcal{W}}^{(k)}$ by
\[
\widetilde{\mathcal{W}}^{(k)}_{S,\tau}
=
\alpha^{(k)}_{S}(\tau),
\]
all interaction components based on STII can be expressed in a unified manner as a tensor contraction with the value tensor. Each entry of the weight tensor $\widetilde{\mathcal{W}}^{(k)}$ is given by
\begin{equation}
\label{eq:widetilde_W}
\widetilde{\mathcal{W}}^{(k)}_{S,\tau}
=
\frac{k}{n_{\mathrm{in}}}
\sum_{\substack{
T\subseteq [n_{\mathrm{in}}]\setminus S,\;
W\subseteq S\\
T\cup W = U(\tau)
}}
\frac{(-1)^{|S|-|W|}}{\binom{n_{\mathrm{in}}-1}{|T|}},
\end{equation}
where the derivation is provided in Appendix~\eqref{app:cal_widetilde_W}.

\begin{theorem}[IT\textnormal{-}SHAP Representation]
\label{thm:IT-SHAP-definition}
Let $\mathcal{S}_{k}=\{\, S\subseteq[n_{\mathrm{in}}]: |S|\le k \,\}$. The IT\textnormal{-}SHAP value tensor of order $k$ is given by
\begin{equation}
\label{eq:it-shap-def}
\mathcal{T}^{(k)}(M,P)
=
\widetilde{\mathcal{W}}^{(k)}
\times_{2}
\mathcal{V}^{(M,P)}
\in
\mathbb{R}^{\,\mathbb{D}\times|\mathcal{S}_{k}|\times n_{\mathrm{out}}},
\end{equation}
where $\widetilde{\mathcal{W}}^{(k)}$ is the STII-based weight tensor and $\mathcal{V}^{(M,P)}$ is the value tensor defined in MST.
\end{theorem}

The representation in \eqref{eq:it-shap-def} shows that STII can be realized as an explicit multilinear map acting on the value tensor. Consequently, IT\textnormal{-}SHAP inherits the axiomatic properties satisfied by STII.

\begin{lemma}[Consistency with MST]
\label{lem:IT-equals-MST}
For $k=1$, IT\textnormal{-}SHAP coincides with MST.
\end{lemma}

\begin{proof}[Proof sketch]
When $k=1$, the weight tensor $\widetilde{\mathcal{W}}^{(1)}$ extracts only first-order differences. The resulting tensor contraction coincides with the definition of MST.
\end{proof}

For reference, when $k=2$ and $S=\{i,j\}$, the second-order discrete derivative is given by
\begin{equation}
\label{eq:second-order-delta}
\delta_{\{i,j\}}F(T)
=
F(T\cup\{i,j\})
-
F(T\cup\{i\})
-
F(T\cup\{j\})
+
F(T),
\end{equation}
and IT\textnormal{-}SHAP realizes the corresponding second-order interaction defined by STII within the same tensor contraction framework.

\subsection{Computational Complexity and Scalability}

This subsection formulates the computational structure of the proposed IT\textnormal{-}SHAP from the perspective of computational complexity and establishes that, under a TT assumption, the computation of IT\textnormal{-}SHAP belongs to the complexity class NC$^{2}$. The discussion proceeds in two stages. First, we show that under a general TN structure, the computation of IT\textnormal{-}SHAP is intrinsically intractable. Second, we demonstrate that when a TT structure is assumed, the weight tensor defining IT\textnormal{-}SHAP admits a TT representation, which implies that the entire computation can be executed by parallel algorithms of polylogarithmic depth.

It is important to emphasize that, among the components constituting IT\textnormal{-}SHAP, the value tensor corresponding to the value function has exactly the same structure as that used in MST in prior work. Therefore, the computability of IT\textnormal{-}SHAP depends essentially not on the value function side, but on the structure of the Modified Weighted Coalitional Tensor newly introduced in this study.

\subsubsection{General TN Structure}

We first consider the case where a general TNs representation is assumed. \cite{marzouk2025shap} showed that the computation of MST is $\#\mathrm{P}$-hard under a general TN structure. This hardness arises from the fact that tensor contraction inherently encodes an exponential combinatorial structure.

The same argument applies to IT\textnormal{-}SHAP introduced in this work. Specifically, as long as the model $M$ and the background distribution $P$ are given as general TNs representation, the computation of IT\textnormal{-}SHAP is also $\#\mathrm{P}$-hard.

\begin{lemma}[Hardness under a General TN Structure]
\label{lem:IT-TN-P-hard}
Assume that the machine learning model $M$ and the background distribution $P$ are given by general TNs representation. Then, the computation of IT\textnormal{-}SHAP is $\#\mathrm{P}$-hard.
\end{lemma}

This result shows that fast exact computation of IT\textnormal{-}SHAP cannot be expected under unrestricted TNs representation.

\subsubsection{TT Structure}

We next consider the case where the model $M$ and the background distribution $P$ are given as tensors in TT form. The TT representation corresponds to a one-dimensional chain TN, and its contraction can be expressed as a sequence of matrix multiplications. Consequently, TT contractions are known to admit parallel algorithms of polylogarithmic depth \cite{marzouk2025shap}.

Whether the computation of IT\textnormal{-}SHAP belongs to NC$^{2}$ reduces to whether the Modified Weighted Coalitional Tensor specific to IT\textnormal{-}SHAP, in addition to the value tensor, admits a TT representation. We show below that for any fixed interaction order $k$, the Modified Weighted Coalitional Tensor can indeed be constructed in TT form.

\begin{lemma}[TT Representability of the Modified Weighted Coalitional Tensor]
\label{lem:IT-TT-MWCT}
Assume that the machine learning model $M$ and the background distribution $P$ are given as tensors in TT form. Then, for any fixed interaction order $k$, the Modified Weighted Coalitional Tensor associated with IT\textnormal{-}SHAP admits a TT representation.
\end{lemma}

\begin{proof}[Proof sketch]
For the case $k=1$, corresponding to MST, it is known that the associated weight tensor admits a TT representation \cite{marzouk2025shap}. For a general order $k$, the Modified Weighted Coalitional Tensor can be expressed, by definition, as a composition of $k$ first-order discrete difference operators. Specifically, for $S=\{i_{1},\ldots,i_{k}\}$,
\[
\widetilde{\mathcal{W}}^{(k)}_{S}
=
\delta_{i_{k}}\cdots\delta_{i_{1}}\,\widetilde{\mathcal{W}}^{(0)}.
\]
Each discrete difference operator $\delta_{i}$ acts as a linear operator on the tensor, affecting only the mode corresponding to the $i$-th feature.

Applying a discrete difference operator $\delta_{i}$ to a tensor represented in TT form preserves the TT structure, and the resulting increase in TT ranks is bounded by a constant factor. Therefore, by induction on $k$, starting from the base case $k=1$, we conclude that $\widetilde{\mathcal{W}}^{(k)}$ admits a TT representation for any fixed $k$.
\end{proof}

Lemma~\ref{lem:IT-TT-MWCT} guarantees that both the weight tensor and the value tensor defining IT\textnormal{-}SHAP admit TT representations. Consequently, the computation of IT\textnormal{-}SHAP can be implemented as a contraction of TT tensors and thus admits a parallel algorithm of polylogarithmic depth. We summarize this consequence in the following theorem.

\begin{theorem}[NC$^{2}$ Computability of IT\textnormal{-}SHAP]
\label{thm:IT-TT-NC2}
Under the assumptions of Lemma~\ref{lem:IT-TT-MWCT}, any interaction component $\mathcal{T}^{(k)}(M,P)$ defined by IT\textnormal{-}SHAP can be computed in the parallel complexity class NC$^{2}$ using a polynomial number of processors.
\end{theorem}

\subsubsection{Upper and Lower Bounds on the Number of Processors}

A natural question is whether the number of processors required to compute IT\textnormal{-}SHAP can be evaluated constructively, in a manner analogous to \cite{marzouk2025shap}. In prior work, explicit TT construction and contraction algorithms were provided for SHAP, and by counting the number of operations executed in parallel at each step, a constructive upper bound on the required number of processors was derived. In particular, it was shown that SHAP can be computed in parallel time $\mathcal{O}(\log^{2} n_{\mathrm{in}})$ using $\mathcal{O}(n_{\mathrm{in}}^{3})$ processors \cite{marzouk2025shap}.

In principle, a similar constructive analysis could be carried out for IT\textnormal{-}SHAP. However, IT\textnormal{-}SHAP outputs all interaction components up to order $k$, whose index set is
\[
\mathcal{S}_{k}
=
\{\, S \subseteq [n_{\mathrm{in}}] \mid |S|\le k \,\}.
\]
The cardinality of this set is
\[
|\mathcal{S}_{k}|
=
\sum_{i=0}^{k} \binom{n_{\mathrm{in}}}{i},
\]
and for fixed $k$,
\[
|\mathcal{S}_{k}| = \Theta\!\left(n_{\mathrm{in}}^{k}\right).
\]
The computation of IT\textnormal{-}SHAP requires producing one interaction value $I^{(k)}_{S}(F)$ for each $S\in\mathcal{S}_{k}$. For a general value function $F$, these interaction values are independent real numbers, and the output corresponding to one subset $S$ does not determine the output for another subset $S'$. Therefore, computing and outputting all $k$-th order interaction components necessarily requires at least
\[
\Omega\!\left(|\mathcal{S}_{k}|\right)
=
\Omega\!\left(n_{\mathrm{in}}^{k}\right)
\]
elementary computation or write operations. This lower bound is information-theoretic in nature and follows solely from the output size of the problem, independent of algorithmic details or contraction order.

We now consider the PRAM model of parallel computation. In this model, the total work $W$, parallel time $T$, and number of processors $P$ satisfy the inequality
\begin{equation}
\label{eq:brent-inequality}
W \le T \times P,
\end{equation}
known as Brent's inequality \cite{brent1974parallel}. In our setting, we fix the parallel time to the NC$^{2}$ regime,
\[
T = O\!\left(\log^{2} n_{\mathrm{in}}\right).
\]
Substituting the information-theoretic lower bound $W \ge \Omega(n_{\mathrm{in}}^{k})$ yields
\[
P
\;\ge\;
\frac{W}{T}
\;\ge\;
\Omega\!\left(
\frac{n_{\mathrm{in}}^{k}}{\log^{2} n_{\mathrm{in}}}
\right).
\]
Here, an information-theoretic lower bound refers to a bound derived solely from the amount of information contained in the input and output, independent of algorithmic structure, implementation details, or computation order \cite{borodin2023time}.

Therefore, although it is in principle possible to derive constructive upper bounds on the number of processors required for IT\textnormal{-}SHAP, the scale of such bounds necessarily grows proportionally to the total number of interaction terms that must be output. For this reason, rather than pursuing refined constructive upper bounds as in \cite{marzouk2025shap}, this study focuses on lower-bound analysis for the number of processors under a fixed parallel depth. This lower-bound perspective allows us to characterize, in an implementation-independent manner, the intrinsic scalability limits of higher-order interaction computation.

\subsection{Structure of Discrete Difference Operators}
\label{subsec:delta-structure}

This subsection shows that the Modified Weighted Coalitional Tensor induced by STII admits a TT structure by analyzing the linear-algebraic properties of discrete difference operators. The central observation is that each higher-order discrete difference operator $\delta_i$ can be represented as a linear map acting on the function space over $\{0,1\}^{n_{\mathrm{in}}}$, and that its action is mode-wise local and low rank. We first characterize the structure of a single operator $\delta_i$, then verify that compositions of such operators preserve the TT structure, and finally conclude that the STII-based Modified Weighted Coalitional Tensor necessarily has a TT representation.

\subsubsection{Linear-algebraic structure of discrete difference operators}

We begin by formulating the first-order discrete difference operator $\delta_i$ acting on set functions as an explicit linear map on the tensor space $\mathbb{R}^{2\times\cdots\times 2}$. This formulation reveals that $\delta_i$ acts locally on a single mode and has rank one.

A set function
\[
F : 2^{[n_{\mathrm{in}}]} \to \mathbb{R}
\]
can be identified with a function on $\{0,1\}^{n_{\mathrm{in}}}$ via the indicator correspondence
\[
S \subseteq [n_{\mathrm{in}}]
\;\longleftrightarrow\;
\mathbf{s}=(s_1,\ldots,s_{n_{\mathrm{in}}}) \in \{0,1\}^{n_{\mathrm{in}}},
\qquad
s_i = \mathbf{1}_{\{i\in S\}}.
\]
Under this identification, $F$ is represented as an $n_{\mathrm{in}}$-th order tensor
\[
\mathcal{F}_{s_1,\ldots,s_{n_{\mathrm{in}}}}
=
F\bigl(\{\, j \mid s_j = 1 \,\}\bigr),
\qquad
\mathcal{F} \in \mathbb{R}^{2 \times \cdots \times 2}.
\]

For a single feature $i\in[n_{\mathrm{in}}]$, the first-order discrete difference operator
\[
\delta_i F(T) = F(T \cup \{i\}) - F(T),
\qquad
T \subseteq [n_{\mathrm{in}}]\setminus\{i\},
\]
can be written as a linear map acting on $\mathcal{F}$. Specifically, letting $I_2$ denote the $2\times 2$ identity matrix, we have
\begin{equation}
\label{eq:delta-as-linear-map-expanded}
\delta_i
=
\underbrace{I_2 \otimes \cdots \otimes I_2}_{i-1}
\;\otimes\;
D
\;\otimes\;
\underbrace{I_2 \otimes \cdots \otimes I_2}_{n_{\mathrm{in}}-i},
\end{equation}
or, equivalently,
\begin{equation}
\label{eq:delta-as-linear-map}
\delta_i
=
I_2^{\otimes (i-1)} \otimes D \otimes I_2^{\otimes (n_{\mathrm{in}}-i)},
\end{equation}
where
\[
D =
\begin{pmatrix}
-1 & 1 \\
0 & 0
\end{pmatrix}.
\]
The component-wise action of $\delta_i$ is therefore given by
\begin{equation}
\label{eq:delta-component}
(\delta_i \mathcal{F})_{s_1,\ldots,s_{n_{\mathrm{in}}}}
=
\sum_{t_i=0}^{1}
D_{s_i,t_i}
\,
\mathcal{F}_{s_1,\ldots,t_i,\ldots,s_{n_{\mathrm{in}}}}.
\end{equation}
By construction of $D$, this yields
\[
(\delta_i \mathcal{F})_{\mathbf{s}}
=
-\mathcal{F}_{s_1,\ldots,0,\ldots,s_{n_{\mathrm{in}}}}
+
\mathcal{F}_{s_1,\ldots,1,\ldots,s_{n_{\mathrm{in}}}}
=
F(T\cup\{i\}) - F(T),
\]
confirming that $\delta_i$ correctly implements the first-order discrete difference.

Equation~\eqref{eq:delta-as-linear-map} makes explicit that $\delta_i$ is a linear operator on $\mathbb{R}^{2\times\cdots\times 2}$. Consequently, a higher-order discrete difference operator
\begin{equation}
\label{eq:Higher-order discrete difference operator}
\delta_S = \prod_{i\in S} \delta_i
\end{equation}
is defined as a product of linear operators. This establishes the fundamental linear-algebraic nature of discrete difference operators.

From \eqref{eq:delta-component}, it is evident that $(\delta_i \mathcal{F})_{s_1,\ldots,s_{n_{\mathrm{in}}}}$ depends only on the index $s_i$ and leaves all other indices unchanged. Hence, $\delta_i$ is a mode-wise local linear operator that does not introduce couplings between different modes.

We next examine its rank structure. The matrix
\[
D =
\begin{pmatrix}
-1 & 1 \\
0 & 0
\end{pmatrix}
\]
satisfies $\mathrm{rank}(D)=1$. Therefore, $\delta_i$ is a rank-$1$ along the $i$-th mode. This can also be seen via matricization. Let $\mathcal{F}_{(k)}\in\mathbb{R}^{2\times 2^{n_{\mathrm{in}}-1}}$ denote the mode-$k$ unfolding. Then
\[
(\delta_i \mathcal{F})_{(k)}
=
\begin{cases}
D\,\mathcal{F}_{(k)}, & i = k, \\[1mm]
\mathcal{F}_{(k)}
\bigl(
I_2^{\otimes(\cdots)} \otimes D \otimes I_2^{\otimes(\cdots)}
\bigr), & i \neq k.
\end{cases}
\]
Since $\mathrm{rank}(D)=1$ and tensor products with identity matrices preserve rank, the induced change in the unfolding has rank at most one. This confirms that $\delta_i$ acts as a rank-$1$ perturbation.

\subsubsection{Action of discrete difference operators on TT tensors}

We now analyze how $\delta_i$ and its composition $\delta_S$ act on tensors represented in TT form. The key result is that $\delta_i$ is absorbed locally into a single TT core without altering the network connectivity.

\begin{theorem}[Preservation of TT structure under rank-$1$ mode-wise operators]
\label{thm:rank1-modewise-TT}
Let $\mathcal{F}\in\mathbb{R}^{2\times\cdots\times 2}$ be represented in TT form. For each $i\in[n_{\mathrm{in}}]$, consider the linear operator
\[
\delta_i
=
I_2^{\otimes(i-1)} \otimes D \otimes I_2^{\otimes(n_{\mathrm{in}}-i)},
\qquad
\mathrm{rank}(D)=1.
\]
Then, the action of $\delta_i$ affects only the TT core corresponding to mode $i$ and preserves the TT structure. Moreover, for any finite set $S\subseteq[n_{\mathrm{in}}]$, the composed operator
\[
\delta_S=\prod_{i\in S}\delta_i
\]
also preserves the TT structure of $\mathcal{F}$.
\end{theorem}

\begin{proof}[Proof sketch]
The operator $\delta_i$ acts as the identity operator on all modes except the $i$-th. Hence, in the TT representation
\[
\mathcal{F}_{s_1,\ldots,s_{n_{\mathrm{in}}}}
=
\sum_{\alpha_0,\ldots,\alpha_{n_{\mathrm{in}}}}
G^{(1)}_{\alpha_0,s_1,\alpha_1}
\cdots
G^{(n_{\mathrm{in}})}_{\alpha_{n_{\mathrm{in}}-1},s_{n_{\mathrm{in}}},\alpha_{n_{\mathrm{in}}}},
\]
the action of $\delta_i$ can be absorbed by replacing the $i$-th core with
\[
\widetilde{G}^{(i)} = D \cdot G^{(i)}.
\]
All other cores remain unchanged, and the TT connectivity is preserved. Since $\delta_i$ and $\delta_j$ commute for $i\neq j$, the composition $\delta_S$ is realized as successive local updates to the corresponding cores. A detailed component-wise derivation is provided in Appendix~\ref{app:proof-rank1-TT}.
\end{proof}

\subsubsection{Failure of Tucker and CP representations under rank-$1$ mode-wise operators}

We next explain why the structure-preserving property in Theorem~\ref{thm:rank1-modewise-TT} is specific to the TT representation and does not extend to other standard tensor representations.

For the Tucker representation, let
\[
\mathcal{F}
=
\mathcal{G}
\times_1 U^{(1)}
\times_2 \cdots
\times_{n_{\mathrm{in}}} U^{(n_{\mathrm{in}})},
\]
where the core tensor $\mathcal{G}$ couples all modes simultaneously. Although a single operator $\delta_i$ can be absorbed as a left multiplication of $U^{(i)}$ by $D$, compositions of operators $\delta_i$ and $\delta_{i'}$ for $i\neq i'$ necessarily interact through the shared core $\mathcal{G}$. As a result, their effects cannot be separated into independent local updates, and the Tucker structure is not preserved under finite compositions of rank-$1$ mode-wise operators.

For the CP representation,
\[
\mathcal{F}
=
\sum_{r=1}^{R}
a^{(1)}_r \otimes a^{(2)}_r \otimes \cdots \otimes a^{(n_{\mathrm{in}})}_r,
\]
a single operator $\delta_i$ acts locally on each rank-$1$ term and preserves the CP form. However, applying a higher-order operator $\delta_S$ generates $2^{|S|}$ rank-$1$ terms per original component due to the inclusion--exclusion structure. Consequently, the number of terms grows exponentially with $|S|$, and neither the representation size nor the computational cost can be controlled polynomially.

Therefore, Tucker representations fail to preserve locality, and CP representations suffer from exponential growth under rank-$1$ mode-wise operators. In contrast, the TT representation uniquely preserves both structure and computational tractability under finite compositions of such operators.

\subsubsection{Modified Weighted Coalitional Tensor induced by discrete difference operators}

We now apply the above results to characterize the structure of the Modified Weighted Coalitional Tensor induced by STII. This tensor is generated by applying higher-order discrete difference operators $\delta_S$ to a zeroth-order base tensor, and by the structure-preserving property established above, its TNs representation is necessarily restricted to a one-dimensional chain, namely a TT structure.

From \eqref{eq:widetilde_W}, each entry of the STII-based Modified Weighted Coalitional Tensor is given by
\[
\widetilde{\mathcal{W}}^{(k)}_{S,\tau}
=
\frac{k}{n_{\mathrm{in}}}
\sum_{\substack{
T\subseteq [n_{\mathrm{in}}]\setminus S,\;
W\subseteq S\\
T\cup W = U(\tau)
}}
\frac{(-1)^{|S|-|W|}}{\binom{n_{\mathrm{in}}-1}{|T|}}.
\]
Using the condition $T\cup W=U(\tau)$, this expression can be rewritten as
\[
\sum_{T\subseteq [n_{\mathrm{in}}]\setminus S}
\frac{1}{\binom{n_{\mathrm{in}}-1}{|T|}}
\sum_{W\subseteq S}
(-1)^{|S|-|W|}
\mathbf{1}_{\{T\cup W = U(\tau)\}}.
\]
The inner sum coincides with the definition of the higher-order discrete difference operator
\[
\delta_S F(T)
=
\sum_{W\subseteq S}
(-1)^{|S|-|W|}
F(T\cup W).
\]

To make this correspondence explicit in tensor form, introduce the zeroth-order base tensor
\[
(\mathcal{E}_0)_{\mathbf{s}}
=
\mathbf{1}_{\{\mathbf{s}=\mathbf{0}\}}
\in \mathbb{R}^{2\times\cdots\times 2}.
\]
Then,
\[
(\delta_S \mathcal{E}_0)_{\mathbf{u}}
=
\sum_{W\subseteq S}
(-1)^{|S|-|W|}
\mathbf{1}_{\{U(\mathbf{u}) = W\}},
\]
and \eqref{eq:widetilde_W} can be rewritten as
\begin{equation}
\label{eq:mwct-deltaS}
\widetilde{\mathcal{W}}^{(k)}_{S,\tau}
=
\frac{k}{n_{\mathrm{in}}}
\sum_{T\subseteq [n_{\mathrm{in}}]\setminus S}
\frac{1}{\binom{n_{\mathrm{in}}-1}{|T|}}
\bigl(
\delta_S \mathcal{E}_0
\bigr)_{U(\tau)}.
\end{equation}

Thus, the Modified Weighted Coalitional Tensor is obtained by applying the higher-order discrete difference operator
\[
\delta_S = \prod_{i\in S}\delta_i
\]
to the base tensor $\mathcal{E}_0$. Since $\mathcal{E}_0$ has TT rank $1$ in every mode, its TT representation is trivial. By Theorem~\ref{thm:rank1-modewise-TT}, the application of $\delta_S$ preserves the TT structure. Consequently, the TNs representing the Modified Weighted Coalitional Tensor remains a one-dimensional chain aligned with the natural ordering of modes. Therefore, the STII-based Modified Weighted Coalitional Tensor necessarily admits a TT representation.

\section{Discussion}

This paper proposed IT\textnormal{-}SHAP as a unified framework for representing higher-order interaction indices based on STII as TT-structured TNs. In this section, we consolidate the theoretical insights obtained in this study, clarify their relationship to existing work, and discuss the limitations of the proposed framework together with directions for future research.

The primary contribution of this work is to establish, on a rigorous theoretical basis, that higher-order interactions defined by STII can be expressed explicitly as linear maps acting on a value tensor, and that the Modified Weighted Coalitional Tensor implementing this map necessarily admits a TT structure. Conventionally, STII has been understood through its combinatorial definition based on discrete derivatives and averaging over all permutations, leading to the widespread belief that its computational cost is intrinsically exponential in high-dimensional settings. This study demonstrates that such computational difficulty is not inherent to the interaction index itself, but instead depends critically on the representation structure under which STII is realized.

Specifically, we showed that while the computation of STII-based interactions is $\#\mathrm{P}$-hard under general TNs representation, it becomes tractable when a one-dimensional chain structure in the form of a TT representation is provided as input. In this setting, STII can be reformulated as a tensor contraction and evaluated within the parallel complexity class NC$^{2}$. Consequently, STII is no longer viewed as a purely combinatorial operation, but rather as a contraction problem on TNs, whose computational structure can be characterized precisely within the framework of parallel computation theory.

Moreover, this study demonstrated that the TT structure of the Modified Weighted Coalitional Tensor is not based on approximate assumptions or empirical design choices. Instead, it follows necessarily from the linear-algebraic properties of the discrete difference operators inherent in the definition of STII. In particular, each first-order discrete difference operator $\delta_i$ can be represented as a mode-wise rank-$1$ linear map acting on the function space over $\{0,1\}^{n_{\mathrm{in}}}$. Higher-order operators $\delta_S$ decompose into products of such operators. As a result, the action of $\delta_S$ preserves the underlying TN connectivity and necessarily maintains the one-dimensional chain structure present in the initial representation. Therefore, the TT structure of the Modified Weighted Coalitional Tensor is a direct consequence of the STII definition itself.

Existing studies on SHAP computation can be broadly classified into approximation-based methods and exact methods under structural assumptions. This work belongs to the latter category. While previous approaches such as MST were restricted to first-order contributions, IT\textnormal{-}SHAP enables the exact treatment of interactions of arbitrary order within a single tensorial framework. In particular, IT\textnormal{-}SHAP coincides with MST when $k=1$ and, for $k \ge 2$, directly inherits the discrete difference structure and axiomatic properties prescribed by STII. In this sense, the proposed framework constitutes a natural and rigorous extension of existing theory.

From the perspective of computational complexity, this study clarifies that the difficulty of higher-order interaction computation is not uniform, but instead bifurcates qualitatively depending on the representation structure. Membership in NC$^{2}$ implies that IT\textnormal{-}SHAP can be evaluated with polylogarithmic parallel depth in the number of input features. At the same time, the total number of interaction terms to be output is
\[
|\mathcal{S}_k|
=
\sum_{i=0}^{k} \binom{n_{\mathrm{in}}}{i}
=
\Theta(n_{\mathrm{in}}^{k}),
\]
which grows polynomially even for fixed $k$. Therefore, any algorithm that computes and outputs all interaction terms must incur at least $\Omega(n_{\mathrm{in}}^{k})$ total work. Using the PRAM relation $W \le T \times P$, achieving a parallel time $T = O(\log^{2} n_{\mathrm{in}})$ requires
\[
P
=
\Omega\!\left(
\frac{n_{\mathrm{in}}^{k}}{\log^{2} n_{\mathrm{in}}}
\right)
\]
processors. This lower bound is information-theoretic in nature and arises solely from the output size, independent of algorithmic design.

Conversely, from the viewpoint of upper bounds, constructive computation of IT\textnormal{-}SHAP under the TT structure enables parallel evaluation with a polynomial number of processors and polylogarithmic depth. Although the processor count necessarily scales with the number of interaction terms, the exponential inefficiency in computational depth inherent in naive STII computation is eliminated. This contrast highlights the essential improvement achieved by the proposed framework.

Taken together, this work provides a clear conceptual separation of the difficulty of higher-order interaction computation into two components: (i) an unavoidable cost determined by the output size, and (ii) a representation-dependent cost governing computational depth. IT\textnormal{-}SHAP addresses the latter by exploiting TNs representation, thereby demonstrating that depth-related intractability can be overcome without altering the underlying interaction definition.

At the same time, the proposed framework assumes that both the value tensor and the weight tensor are available in TT form with practically manageable ranks. For general high-dimensional tensors, TT ranks may grow exponentially with the number of modes, making the construction and maintenance of TT representations a potential computational bottleneck. Therefore, practical application to real-world data requires careful consideration of the preprocessing stage, namely whether suitable TT representations can be constructed and maintained within realistic computational resources.

In this regard, recent advances in approximate and accelerated TT decomposition techniques offer promising directions for extending applicability. Examples include fast approximate TT-SVD based on randomized SVD \cite{huber2018randomized}, low-rank approximations using TensorSketch \cite{chen2023low}, and acceleration of TT-ALS via leverage score sampling \cite{bharadwaj2024efficient}. Integrating such techniques with IT\textnormal{-}SHAP remains an important avenue for future work.

In summary, this study reformulated STII-based higher-order interaction analysis from a combinatorial definition into a tensor-analytic framework and provided a rigorous characterization of its computational structure and complexity. The proposed IT\textnormal{-}SHAP framework offers an explicit and unified representation for Shapley theory with higher-order interactions within the contexts of tensor analysis and parallel computation theory.

\section{Conclusion}
This paper provided a rigorous tensor-analytic reformulation of the STII and established a precise characterization of its computational structure through the lens of TNs representation. By interpreting STII as a linear functional acting on a value tensor defined over the Boolean hypercube, we demonstrated that higher-order interaction indices can be realized explicitly as tensor contractions, rather than as inherently combinatorial objects.

The central theoretical result is that the Modified Weighted Coalitional Tensor induced by STII necessarily admits an exact TT representation for any fixed interaction order. This property follows directly from the linear-algebraic structure of discrete finite-difference operators: each first-order operator acts as a rank-$1$, mode-wise local linear map, and finite compositions of such operators preserve the one-dimensional chain structure characteristic of the TT representation. As a consequence, the TT structure of the STII weight tensor is not an approximation or modeling assumption, but an unavoidable implication of the STII definition itself.

Building on this structural result, we established a sharp complexity separation. Under unrestricted TNs representation, the exact computation of STII-based interaction components is \#P-hard, confirming the intrinsic intractability of the problem in the absence of structural constraints. In contrast, when both the value tensor and the STII weight tensor are given in TT form, the evaluation of each interaction component can be performed by parallel algorithms of polylogarithmic depth, placing the computation in the parallel complexity class $\mathrm{NC}^2$. This shows that the exponential difficulty traditionally associated with higher-order interaction analysis arises not from the interaction index itself, but from the representational framework used to implement it.

From a broader perspective, these results clarify that the computational behavior of higher-order linear functionals on set functions is governed by representation structure rather than by combinatorial definition. The proposed IT\textnormal{-}SHAP framework unifies STII and prior tensor-based SHAP formulations within a single algebraic setting, extending exact, parallelizable computation from first-order attributions to interactions of arbitrary order. In doing so, it provides a principled foundation for scalable higher-order explainability in high-dimensional models.

At the same time, the framework relies on the availability of TT representations with manageable ranks. While this assumption is natural in many structured or compressible settings, TT ranks may grow prohibitively in worst-case scenarios. Addressing rank growth, developing guarantees for approximate TT constructions, and integrating randomized or sketching-based tensor techniques remain important directions for future research.

In conclusion, this work reframes higher-order Shapley interaction analysis as a problem in tensor linear algebra and parallel computation. By disentangling intrinsic output-size complexity from representation-dependent computational depth, it establishes that higher-order interaction indices are not fundamentally intractable, but become efficiently computable once their algebraic structure is made explicit through appropriate tensor representations.

\section*{Acknowledgments}
This work was supported by JSPS KAKENHI Grant Number 23K22166.

\section*{Appendix}

\subsection{Derivation of Equation~\ref{eq:widetilde_W}}
\label{app:cal_widetilde_W}

This subsection derives the component-wise representation \eqref{eq:widetilde_W} of the weight tensor $\widetilde{\mathcal{W}}^{(k)}$ from the definition of STII. Throughout this derivation, we assume $|S|=k$. By the definition of STII, we have
\begin{equation}
\label{eq:app_stii_def}
I_S^{(k)}(F)
=
\mathbb{E}_{\pi\in\mathfrak{S}_{n_{\mathrm{in}}}}
\bigl[
\delta_S F(\pi_S)
\bigr].
\end{equation}
Using the definition of the discrete difference operator,
\[
\delta_S F(T)
=
\sum_{W\subseteq S}
(-1)^{|S|-|W|}
F(T\cup W),
\]
we rewrite \eqref{eq:app_stii_def} as
\begin{equation}
\label{eq:app_diff_expand}
I_S^{(k)}(F)
=
\sum_{W\subseteq S}
(-1)^{|S|-|W|}
\mathbb{E}_{\pi}
\bigl[
F(\pi_S \cup W)
\bigr].
\end{equation}

Since $\pi_S \subseteq [n_{\mathrm{in}}]\setminus S$, the expectation over permutations can be expressed as
\begin{equation}
\label{eq:app_perm_average}
\mathbb{E}_{\pi}
\bigl[
F(\pi_S \cup W)
\bigr]
=
\sum_{T\subseteq[n_{\mathrm{in}}]\setminus S}
\frac{1}{n_{\mathrm{in}}}
\frac{1}{\binom{n_{\mathrm{in}}-1}{|T|}}
F(T\cup W).
\end{equation}
Substituting \eqref{eq:app_perm_average} into \eqref{eq:app_diff_expand}, we obtain
\begin{equation}
\label{eq:app_double_sum}
I_S^{(k)}(F)
=
\frac{1}{n_{\mathrm{in}}}
\sum_{W\subseteq S}
\sum_{T\subseteq[n_{\mathrm{in}}]\setminus S}
\frac{(-1)^{|S|-|W|}}{\binom{n_{\mathrm{in}}-1}{|T|}}
F(T\cup W).
\end{equation}

Next, for a routing index $\tau\in\{1,2\}^{n_{\mathrm{in}}}$, define $U(\tau)=\{i:\tau_i=1\}$. Under this notation, the identity $F(T\cup W)=V_{x,\tau,y}$ holds if and only if $U(\tau)=T\cup W$. Therefore, \eqref{eq:app_double_sum} can be rewritten as
\begin{equation}
\label{eq:app_tau_form}
I_S^{(k)}(F)
=
\sum_{\tau\in\{1,2\}^{n_{\mathrm{in}}}}
\left(
\frac{1}{n_{\mathrm{in}}}
\sum_{\substack{
T\subseteq[n_{\mathrm{in}}]\setminus S,\;
W\subseteq S\\
T\cup W = U(\tau)
}}
\frac{(-1)^{|S|-|W|}}{\binom{n_{\mathrm{in}}-1}{|T|}}
\right)
V_{x,\tau,y}.
\end{equation}
Comparing this expression with
\[
I_S^{(k)}(F)
=
\sum_{\tau}
\widetilde{\mathcal{W}}^{(k)}_{S,\tau}
\,V_{x,\tau,y},
\]
we identify
\begin{equation}
\label{eq:app_weight_raw}
\widetilde{\mathcal{W}}^{(k)}_{S,\tau}
=
\frac{1}{n_{\mathrm{in}}}
\sum_{\substack{
T\subseteq[n_{\mathrm{in}}]\setminus S,\;
W\subseteq S\\
T\cup W = U(\tau)
}}
\frac{(-1)^{|S|-|W|}}{\binom{n_{\mathrm{in}}-1}{|T|}}.
\end{equation}
Finally, applying the STII normalization that multiplies the $|S|=k$ component by $k$ yields \eqref{eq:widetilde_W}.

\subsection{Proof of Lemma~\ref{lem:IT-equals-MST}}

\begin{proof}
By definition of IT\textnormal{-}SHAP, the order-$k=1$ IT\textnormal{-}SHAP value tensor is given by
\[
\mathcal{T}^{(1)}(M,P)
=
\widetilde{\mathcal{W}}^{(1)} \times_{2} \mathcal{V}^{(M,P)}
\in
\mathbb{R}^{\,\mathbb{D}\times n_{\mathrm{in}}\times n_{\mathrm{out}}}.
\]
The tensor contraction along the second mode yields, for each input $\mathbf{x}\in\mathcal{X}$, feature $i\in[n_{\mathrm{in}}]$, and output component $y$,
\[
\mathcal{T}^{(1)}(M,P)_{\mathbf{x},i,y}
=
\sum_{\boldsymbol{\tau}\in\{1,2\}^{n_{\mathrm{in}}}}
\widetilde{\mathcal{W}}^{(1)}_{i,\boldsymbol{\tau}}\,
\mathcal{V}^{(M,P)}_{\mathbf{x},\boldsymbol{\tau},y}.
\]

For $k=1$, the corresponding first-order discrete difference operator is
\begin{equation}
\label{eq:first-order-delta}
\delta_{\{i\}}F(T)
=
F(T\cup\{i\})-F(T),
\qquad
T\subseteq[n_{\mathrm{in}}]\setminus\{i\}.
\end{equation}
Substituting this expression into the definition of STII yields the first-order interaction
\[
I^{(1)}_{\{i\}}(F)
=
\sum_{T\subseteq[n_{\mathrm{in}}]\setminus\{i\}}
W(T)\bigl(F(T\cup\{i\})-F(T)\bigr),
\]
where the Shapley weight is given by
\begin{equation}
\label{eq:shapley-weight}
W(T)
=
\frac{|T|!\,(n_{\mathrm{in}}-|T|-1)!}{n_{\mathrm{in}}!}.
\end{equation}

In the MST formulation, a routing index $\boldsymbol{\tau}\in\{1,2\}^{n_{\mathrm{in}}}$ is associated with a subset
\[
T(\boldsymbol{\tau})
=
\{\,j\in[n_{\mathrm{in}}]\setminus\{i\} : \tau_j=1\,\}.
\]
Under this correspondence, each component of the value tensor satisfies
\[
\mathcal{V}^{(M,P)}_{\mathbf{x},\boldsymbol{\tau},y}
=
F_{\mathbf{x},y}\!\left(T(\boldsymbol{\tau})\right).
\]

The weight tensor $\widetilde{\mathcal{W}}^{(1)}$ encodes the signed structure of the first-order discrete difference operator. By construction,
\[
\widetilde{\mathcal{W}}^{(1)}_{i,\boldsymbol{\tau}}
=
\begin{cases}
+\,W\!\left(T(\boldsymbol{\tau})\right), & \text{if }\tau_i=1,\\[4pt]
-\,W\!\left(T(\boldsymbol{\tau})\right), & \text{if }\tau_i=2.
\end{cases}
\]

Using \eqref{eq:shapley-weight} and \eqref{eq:first-order-delta}, we obtain
\[
\begin{aligned}
\mathcal{T}^{(1)}(M,P)_{\mathbf{x},i,y}
&=
\sum_{\boldsymbol{\tau}\in\{1,2\}^{n_{\mathrm{in}}}}
\widetilde{\mathcal{W}}^{(1)}_{i,\boldsymbol{\tau}}\,
\mathcal{V}^{(M,P)}_{\mathbf{x},\boldsymbol{\tau},y} \\
&=
\sum_{T\subseteq[n_{\mathrm{in}}]\setminus\{i\}}
W(T)\bigl(F_{\mathbf{x},y}(T\cup\{i\})-F_{\mathbf{x},y}(T)\bigr).
\end{aligned}
\]
The right-hand side coincides exactly with the definition of the MST contribution of feature $i$,
\[
\mathcal{T}^{\mathrm{MST}}_{\mathbf{x},i,y}
=
\sum_{T\subseteq[n_{\mathrm{in}}]\setminus\{i\}}
W(T)\bigl(F_{\mathbf{x},y}(T\cup\{i\})-F_{\mathbf{x},y}(T)\bigr).
\]
Therefore, IT\textnormal{-}SHAP coincides with MST in the case $k=1$.
\end{proof}

\subsection{Proof of Lemma~\ref{lem:IT-TN-P-hard}}

\begin{proof}
This proof establishes that, for any fixed integer $k \ge 1$, the computational problem $\mathrm{IT\textnormal{-}SHAP}(k)$ is $\#\mathrm{P}$-hard. To this end, we construct a polynomial-time Turing reduction from $\mathrm{IT\textnormal{-}SHAP}(1)$ to $\mathrm{IT\textnormal{-}SHAP}(k)$.

By Theorem~\ref{thm:IT-SHAP-definition}, the order-$k$ IT\textnormal{-}SHAP value tensor
\[
\mathcal{T}^{(k)}(M,P)\in\mathbb{R}^{\,\mathbb{D}\times|\mathcal{S}_k|\times n_{\mathrm{out}}}
\]
is indexed along its second mode by
\[
\mathcal{S}_k=\{\,S\subseteq [n_{\mathrm{in}}] : |S|\le k\,\}.
\]
Since $k$ is fixed, we have
\[
|\mathcal{S}_k|
=
\sum_{j=0}^{k}\binom{n_{\mathrm{in}}}{j}
=
\mathcal{O}(n_{\mathrm{in}}^{k}),
\]
and therefore the enumeration of $\mathcal{S}_k$ and the identification of the index corresponding to each $S\in\mathcal{S}_k$ can be performed in polynomial time with respect to $n_{\mathrm{in}}$.

In particular, for any feature index $i\in[n_{\mathrm{in}}]$, the singleton set $\{i\}$ is contained in $\mathcal{S}_k$ for all $k\ge 1$. Let $s(i)$ denote the index along the second mode corresponding to $\{i\}$. Then, by direct component access to $\mathcal{T}^{(k)}(M,P)$, we can recover the first-order component via
\[
\mathcal{T}^{(1)}(M,P)_{\mathbf{x},i,y}
:=
\mathcal{T}^{(k)}(M,P)_{\mathbf{x},\,s(i),\,y}.
\]
This operation consists solely of index translation and tensor entry access and is therefore computable in polynomial time.

We now address computational hardness. By Lemma~\ref{lem:IT-equals-MST}, IT\textnormal{-}SHAP coincides componentwise with MST in the case $k=1$. Moreover, \cite{marzouk2025shap} show that, when the machine learning model $M$ and background distribution $P$ are given by general TNs representation, the computation of MST is $\#\mathrm{P}$-hard. Consequently,
\[
\mathrm{IT\textnormal{-}SHAP}(1)\text{ is }\#\mathrm{P}\text{-hard}.
\]

Combining this hardness result with the polynomial-time extraction of first-order components described above yields
\[
\mathrm{IT\textnormal{-}SHAP}(1)\;\le_p\;\mathrm{IT\textnormal{-}SHAP}(k)
\]
for any fixed $k\ge 1$, where $A\le_p B$ denotes that problem $A$ is solvable using a single oracle call to problem $B$ together with polynomial-time preprocessing and postprocessing.

Therefore, under general TNs representation, $\mathrm{IT\textnormal{-}SHAP}(k)$ is $\#\mathrm{P}$-hard for any fixed integer $k\ge 1$.
\end{proof}

\subsection{Proof of Lemma~\ref{lem:IT-TT-MWCT}}

\begin{proof}
We first consider the case $k=1$. As shown by \cite{marzouk2025shap}, the first-order weight tensor $\widetilde{\mathcal{W}}^{(1)}$, corresponding to MST, admits an exact TT representation. This serves as the base case of the induction.

Next, fix an arbitrary integer $k\ge 1$ and consider the weight tensor $\widetilde{\mathcal{W}}^{(k)}$. By definition, for any subset $S=\{i_1,\ldots,i_k\}\subseteq[n_{\mathrm{in}}]$, the order-$k$ weight tensor component can be written as the composition of $k$ discrete difference operators applied to the zeroth-order weight tensor $\widetilde{\mathcal{W}}^{(0)}$:
\begin{equation}
\label{eq:mwct-delta-composition}
\widetilde{\mathcal{W}}^{(k)}_S
=
\delta_{i_k}\cdots\delta_{i_1}\,\widetilde{\mathcal{W}}^{(0)}.
\end{equation}

Each operator $\delta_i$ acts exclusively on the $i$th mode of the tensor. Specifically, for any tensor $\mathcal{A}\in\mathbb{R}^{I_1\times\cdots\times I_N}$,
\[
(\delta_i\mathcal{A})_{j_1,\ldots,j_N}
=
\sum_{\ell\in\{1,2\}} c_\ell\,
\mathcal{A}_{j_1,\ldots,\ell,\ldots,j_N},
\]
where the coefficients $c_\ell$ are fixed scalars. Thus, $\delta_i$ can be expressed as a finite-dimensional linear map acting along the $i$th mode, equivalently represented as a mode-wise tensor–matrix product.

For tensors represented in TT representation, the application of a linear map along a single mode corresponds to applying the same linear transformation to the associated TT core, while leaving all other cores unchanged. Consequently, the class of TT tensors is closed under the action of each $\delta_i$.

We now proceed by mathematical induction. Assume that, for a fixed $k\ge 1$, the weight tensor component $\widetilde{\mathcal{W}}^{(k)}_S$ admits a TT representation for every subset $S=\{i_1,\ldots,i_k\}$. Consider an arbitrary $(k+1)$-element subset $S'=\{i_1,\ldots,i_k,i_{k+1}\}$. By definition,
\[
\widetilde{\mathcal{W}}^{(k+1)}_{S'}
=
\delta_{i_{k+1}}\,\widetilde{\mathcal{W}}^{(k)}_{S}.
\]
By the induction hypothesis, $\widetilde{\mathcal{W}}^{(k)}_{S}$ is representable in TT representation, and by the closure property established above, the application of $\delta_{i_{k+1}}$ preserves the TT structure. Hence, $\widetilde{\mathcal{W}}^{(k+1)}_{S'}$ also admits a TT representation.

Together with the base case $k=1$, this completes the induction and establishes that, for any fixed integer $k\ge 1$, the weight tensor $\widetilde{\mathcal{W}}^{(k)}$ is representable in TT representation.
\end{proof}

\subsection{Proof of Theorem~\ref{thm:IT-TT-NC2}}

\begin{proof}
By Lemma~\ref{lem:IT-TT-MWCT}, the weight tensor $\widetilde{\mathcal{W}}^{(k)}$ associated with order-$k$ IT\textnormal{-}SHAP admits a TT representation. By assumption, the value tensor $\mathcal{V}^{(M,P)}$ corresponding to the machine learning model $M$ and background distribution $P$ is also provided in TT representation. Therefore, by definition,
\[
\mathcal{T}^{(k)}(M,P)
=
\widetilde{\mathcal{W}}^{(k)} \times_{2} \mathcal{V}^{(M,P)}
\]
can be expressed as a tensor contraction between two TT tensors.

In TT format, each tensor is represented as a sequence of TT cores
\[
G^{(1)},G^{(2)},\ldots,G^{(m)},
\]
where each core has the form
\[
G^{(i)} \in \mathbb{R}^{r_{i-1}\times d_i \times r_i},
\]
with $r_i$ denoting the TT rank and $d_i$ the dimension of the $i$th mode. The contraction of two TT tensors can be decomposed into local contractions between corresponding TT cores, followed by a sequential combination of the resulting matrices. This procedure can be represented as a chain of finite-dimensional matrix multiplications \cite{marzouk2025shap}.

More precisely, any fixed entry $\mathcal{T}^{(k)}(M,P)_{x,S,y}$ can be written as a specific component of the matrix product
\[
A_1 A_2 \cdots A_m,
\]
where each matrix $A_i$ is obtained by contracting the $i$th TT core of $\widetilde{\mathcal{W}}^{(k)}$ with the corresponding TT core of $\mathcal{V}^{(M,P)}$. The matrix dimensions satisfy
\[
A_i \in \mathbb{R}^{\left(r^{(W)}_{i-1} r^{(V)}_{i-1}\right)\times\left(r^{(W)}_{i} r^{(V)}_{i}\right)},
\]
where $r^{(W)}_i$ and $r^{(V)}_i$ denote the intermediate TT ranks of $\widetilde{\mathcal{W}}^{(k)}$ and $\mathcal{V}^{(M,P)}$, respectively.

The matrix product chain $A_1 A_2 \cdots A_m$ can be evaluated using a balanced binary tree scheme, yielding a parallel circuit of depth $\mathcal{O}(\log m)$. Since each individual matrix multiplication involves matrices of size polynomial in $n_{\mathrm{in}}$, standard parallel matrix multiplication algorithms compute each product with depth $\mathcal{O}(\log n_{\mathrm{in}})$. Consequently, the total parallel depth required to compute a fixed component $\mathcal{T}^{(k)}(M,P)_{x,S,y}$ is $\mathcal{O}(\log m \cdot \log n_{\mathrm{in}})$. Using $m=\Theta(n_{\mathrm{in}})$, this simplifies to $\mathcal{O}(\log^{2} n_{\mathrm{in}})$.

Therefore, each fixed component of IT\textnormal{-}SHAP can be evaluated using a polynomial number of processors within parallel time $\mathcal{O}(\log^{2} n_{\mathrm{in}})$. It follows that the IT\textnormal{-}SHAP computation belongs to the complexity class NC$^{2}$.
\end{proof}

\subsection{Proof of Theorem~\ref{thm:rank1-modewise-TT}}
\label{app:proof-rank1-TT}
\begin{proof}
Let $\mathcal F \in \mathbb R^{2\times\cdots\times 2}$ be a $n_{\mathrm{in}}$th-order tensor, and suppose that $\mathcal F$ admits the following TT representation:
\[
\mathcal F_{s_1,\ldots,s_{n_{\mathrm{in}}}}
=
\sum_{\alpha_0,\ldots,\alpha_{n_{\mathrm{in}}}}
G^{(1)}_{\alpha_0,s_1,\alpha_1}
G^{(2)}_{\alpha_1,s_2,\alpha_2}
\cdots
G^{(n_{\mathrm{in}})}_{\alpha_{n_{\mathrm{in}}-1},s_{n_{\mathrm{in}}},\alpha_{n_{\mathrm{in}}}},
\]
where $\alpha_0=\alpha_{n_{\mathrm{in}}}=1$, and each TT core satisfies
\[
G^{(i)} \in \mathbb R^{r_{i-1}\times 2\times r_i}.
\]

For each $i\in[n_{\mathrm{in}}]$, define the discrete difference operator $\delta_i$ by
\[
\delta_i
=
I_2^{\otimes(i-1)} \otimes D \otimes I_2^{\otimes(n_{\mathrm{in}}-i)},
\qquad
D=
\begin{pmatrix}
-1 & 1\\
0 & 0
\end{pmatrix}.
\]
The action of $\delta_i$ on the tensor $\mathcal F$ is given componentwise by
\[
(\delta_i \mathcal F)_{s_1,\ldots,s_{n_{\mathrm{in}}}}
=
\sum_{t_i=0}^1
D_{s_i,t_i}
\,
\mathcal F_{s_1,\ldots,t_i,\ldots,s_{n_{\mathrm{in}}}}.
\]

Substituting the TT representation of $\mathcal F$ yields
\[
(\delta_i \mathcal F)_{s_1,\ldots,s_{n_{\mathrm{in}}}}
=
\sum_{t_i=0}^1
D_{s_i,t_i}
\sum_{\alpha_0,\ldots,\alpha_{n_{\mathrm{in}}}}
\prod_{j=1}^{n_{\mathrm{in}}}
G^{(j)}_{\alpha_{j-1},\xi_j,\alpha_j},
\]
where
\[
\xi_j=
\begin{cases}
t_i, & j=i,\\
s_j, & j\neq i.
\end{cases}
\]

Since all sums are finite, the order of summation can be exchanged, giving
\[
(\delta_i \mathcal F)_{s_1,\ldots,s_{n_{\mathrm{in}}}}
=
\sum_{\alpha_0,\ldots,\alpha_{n_{\mathrm{in}}}}
\left(
\prod_{j\neq i}
G^{(j)}_{\alpha_{j-1},s_j,\alpha_j}
\right)
\sum_{t_i=0}^1
D_{s_i,t_i}
G^{(i)}_{\alpha_{i-1},t_i,\alpha_i}.
\]

Define a new TT core associated with the $i$th mode by
\[
\widetilde G^{(i)}_{\alpha_{i-1},s_i,\alpha_i}
:=
\sum_{t_i=0}^1
D_{s_i,t_i}
G^{(i)}_{\alpha_{i-1},t_i,\alpha_i}.
\]
By construction, this satisfies
\[
\widetilde G^{(i)} = D \cdot G^{(i)},
\]
and hence $\widetilde G^{(i)} \in \mathbb R^{r_{i-1}\times 2\times r_i}$.

Using this definition, we obtain
\[
(\delta_i \mathcal F)_{s_1,\ldots,s_{n_{\mathrm{in}}}}
=
\sum_{\alpha_0,\ldots,\alpha_{n_{\mathrm{in}}}}
G^{(1)}_{\alpha_0,s_1,\alpha_1}
\cdots
\widetilde G^{(i)}_{\alpha_{i-1},s_i,\alpha_i}
\cdots
G^{(n_{\mathrm{in}})}_{\alpha_{n_{\mathrm{in}}-1},s_{n_{\mathrm{in}}},\alpha_{n_{\mathrm{in}}}}.
\]

Therefore,
\[
\delta_i \mathcal F
=
G^{(1)} \bullet \cdots \bullet \widetilde G^{(i)} \bullet \cdots \bullet G^{(n_{\mathrm{in}})},
\]
which shows that the action of $\delta_i$ is localized entirely to the $i$th TT core and does not alter any other TT core or the underlying contraction structure.

Next, consider a finite subset $S\subseteq[n_{\mathrm{in}}]$ and define
\[
\delta_S=\prod_{i\in S}\delta_i.
\]
Since $\delta_i\delta_j=\delta_j\delta_i$ for all $i\neq j$, the operator $\delta_S$ is well defined as the successive application of the individual $\delta_i$. By the argument above, each $\delta_i$ acts by replacing the corresponding TT core $G^{(i)}$ with
\[
G^{(i)} \longmapsto D\cdot G^{(i)}.
\]
Consequently, $\delta_S \mathcal F$ admits the TT representation
\[
\delta_S \mathcal F
=
\widetilde G^{(1)} \bullet \widetilde G^{(2)} \bullet \cdots \bullet \widetilde G^{(n_{\mathrm{in}})},
\qquad
\widetilde G^{(i)}=
\begin{cases}
D\cdot G^{(i)}, & i\in S,\\
G^{(i)}, & i\notin S.
\end{cases}
\]

Each update is a local linear transformation applied to an existing TT core and introduces neither new internal indices nor additional contraction edges. Hence, after the action of $\delta_S$, the TN connectivity remains a one-dimensional chain.

In summary, rank-$1$ mode-wise linear operators $\delta_i$, as well as their finite compositions $\delta_S$, act on a TT-represented tensor $\mathcal F$ by updating only the corresponding TT cores while preserving the underlying TN structure. This establishes Theorem~\ref{thm:rank1-modewise-TT} rigorously at the component level.
\end{proof}

\bibliographystyle{siamplain}
\bibliography{references}
\end{document}

%% file: ex_shared.tex
\usepackage{lipsum}
\usepackage{amsfonts}
\usepackage{amssymb}
\usepackage{graphicx}
\usepackage{epstopdf}
\usepackage{algorithmic}
\ifpdf
  \DeclareGraphicsExtensions{.eps,.pdf,.png,.jpg}
\else
  \DeclareGraphicsExtensions{.eps}
\fi

\newsiamremark{remark}{Remark}
\newsiamremark{hypothesis}{Hypothesis}
\crefname{hypothesis}{Hypothesis}{Hypotheses}
\newsiamthm{claim}{Claim}

\headers{Interaction Tensor SHAP}{H. Hasegawa, and Y. Okada}

\title{Interaction Tensor SHAP\thanks{Submitted to the editors DATE.
\funding{This work was supported by JSPS KAKENHI Grant Number 23K22166.}}}

\author{Hiroki Hasegawa\thanks{Graduate School of Science and Technology, University of Tsukuba, Tsukuba, Ibaraki, Japan 
  (\email{hasegawa.hiroki.tkb\_en@u.tsukuba.ac.jp}).}
\and Yukihiko Okada\thanks{Institute of Systems and Information Engineering, Tsukuba Institute for Advanced Research, Center for Artificial Intelligence Research, University of Tsukuba, Tsukuba, Ibaraki, Japan 
  (\email{okayu@sk.tsukuba.ac.jp}).}}

\usepackage{amsopn}

\makeatletter
\newcommand*{\addFileDependency}[1]{% argument=file name and extension
  \typeout{(#1)}% latexmk will find this if $recorder=0 (however, in that case, it will ignore #1 if it is a .aux or .pdf file etc and it exists! if it doesn't exist, it will appear in the list of dependents regardless)
  \@addtofilelist{#1}% if you want it to appear in \listfiles, not really necessary and latexmk doesn't use this
  \IfFileExists{#1}{}{\typeout{No file #1.}}% latexmk will find this message if #1 doesn't exist (yet)
}
\makeatother

\newcommand*{\myexternaldocument}[1]{%
    \externaldocument{#1}%
    \addFileDependency{#1.tex}%
    \addFileDependency{#1.aux}%
}